\RequirePackage{fix-cm}
\documentclass[smallextended]{svjour3}       % onecolumn (second format)
\smartqed  % flush right qed marks, e.g. at end of proof
\usepackage{graphicx}
\usepackage{amssymb,amsmath,epsfig}
\usepackage{algorithm}
\usepackage{algorithmic}
\usepackage{booktabs}
\usepackage{color}
\usepackage{textcomp}
\usepackage{bbding}
\usepackage{pifont}
\usepackage{wasysym}
\usepackage{amssymb}
\usepackage{subcaption}
\usepackage{multirow}
\usepackage{cite}
\usepackage[utf8]{inputenc}
\begin{document}

\title{Leaning Compact and Representative Features for Cross-Modality Person Re-Identification%\thanks{Grants or other notes
}

\titlerunning{Leaning Compact and Representative Features}        % if too long for running head

\author{Guangwei Gao$^{1,2}$\textsuperscript{$\dagger$} \and Hao Shao$^{1}$\textsuperscript{$\dagger$}\thanks{$\dagger$Equal contributions.}   \and Fei Wu$^{1}$  \and Meng Yang$^{3}$ \and Yi Yu$^{2}$ %etc.
}

\authorrunning{Guangwei Gao et al.} % if too long for running head

\institute{Guangwei Gao \at
              csggao@gmail.com           %  \\
           \and
            Hao Shao \at
            sh\_0307@163.com
            \and
             Fei Wu \at
            wufei\_8888@126.com
            \and
             Meng Yang \at
            yangmengpolyu@gmail.com
            \and
            Yi Yu \at
            yiyu@nii.ac.jp
            \and
            $^{1}$ Institute of Advanced Technology, Nanjing University of Posts and Telecommunications, Nanjing, China \at 
            \and
            $^{2}$ Digital Content and Media Sciences Research Division, National Institute of Informatics, Tokyo, Japan \at
            \and
            $^{3}$ Key Laboratory of Machine Intelligence and Advanced Computing, Ministry of Education, Sun Yat-sen University, Guangzhou, China \\
}

\date{Received: date / Accepted: date}

\maketitle

\begin{abstract}
This paper pays close attention to the cross-modality visible-infrared person re-identification (VI Re-ID) task, which aims to match pedestrian samples between visible and infrared modes. In order to reduce the modality-discrepancy between samples from different cameras, most existing works usually use constraints based on Euclidean metric. Because of the Euclidean based distance metric strategy cannot effectively measure the internal angles between the embedded vectors, the existing solutions cannot learn the angularly discriminative feature embedding. Since the most important factor affecting the classification task based on embedding vector is whether there is an angularly discriminative feature space, in this paper, we present a new loss function called Enumerate Angular Triplet (EAT) loss. Also, motivated by the knowledge distillation, to narrow down the features between different modalities before feature embedding, we further present a novel Cross-Modality Knowledge Distillation (CMKD) loss. Benefit from the above two considerations, the embedded features are discriminative enough in a way to tackle modality-discrepancy problem. The experimental results on RegDB and SYSU-MM01 datasets have demonstrated that the proposed method is superior to the other most advanced methods in terms of impressive performance. Code is available at https://github.com/IVIPLab/LCCRF.
\end{abstract}

\section{Introduction}
\label{sec1}

Image retrieval is a research hotspot in computer vision community~\cite{lu2020deep}. Among them, person re-identification refers to matching pedestrian images acquired from disjoint cameras~\cite{8640834,gao2017learning,gao2020hierarchical}. In recent years, it has received substantial attention due to its significant practical value in video surveillance~\cite{RN149,li2019deep}. Conventional person re-identification is only devoted to single-modality,i.e. all the person images are taken by visible cameras during daytime. Nevertheless, the visible cameras cannot image clearly in the dark environment, which impedes the popularization and application of person re-identification~\cite{RN181}. To overcome this obstacle, in addition to the visible cameras, infrared cameras that are robust to illumination variants are also equipped in many surveillance scenarios. Therefore, in practice, we often need to match visible (RGB) and Infrared (IR) pedestrian images.

To narrow the modality gap between the infrared and visible images, recent works~\cite{RN48,2018Cross} used Euclidean metric-based constraints~\cite{RN190,RN191} to force the features from the same identity to be closer than those from the different identities. Although the above approaches have been successful, there still exists an inherent defect in the design of the loss function based on the Euclidean measure: the angle between embedded features cannot be effectively constrained by triple loss, which leads to the indistinguishable direction of features in common space. In the training stage, the included angles of the negative sample pairs may be smaller than that of the positive sample pairs, which makes the model impossible to divide the appropriate area for features to be embedded in the common space. Therefore, the angularly discriminative feature space that we expect is often not available.

More importantly, an angularly discriminative feature space is crucial to the final classification loss in the liner layer. It is  calculating the dot product between feature vectors and weight vectors. When the weight vectors are assured, then the result is determined by the included angles of the feature vectors only. To solve the above problems, we design a loss function named Enumerate Angular Triplet (EAT) loss, which focuses on the included Angle between the embedded vectors generated by different modes and uses the cosine distance to measure the included Angle between the embedded vectors. Most previous methods first learned the unique features of the modal in the feature extraction stage, then mapped the features between different modalities into a common space and narrowed the distance between them. But they overlooked an important issue, the insurmountable gap between the different modalities still exists. Therefore, in the feature extraction stage, extracting the unique features of the modal while reducing the distance between them is authentically beneficial to the subsequent feature embedding stage. To this end, we also propose a new loss function, named Cross-Modality Knowledge Distillation (CMKD) loss, to narrow the distance between different modal features at the end of the unique feature extraction stage.

Our main contributions can be listed as three-fold:
\begin{itemize}
\item We devise an efficient Enumerate Angular Triplet (EAT) loss, which can better help to obtain an angularly separable common feature space via explicitly restraining the internal angles between different embedding features, contributing to the improvement of the performance. 
\item Motivated by the knowledge distillation, a novel   Cross-Modality Knowledge Distillation (CMKD) loss is  proposed to reduce the modality discrepancy in the  modality-specific feature extraction stage, contributing to the effectiveness of the cross-modality person Re-ID task.  
\item Our network achieves prominent results on both SYSU-MM01 and RegDB datasets without any other data augment skills. It achieves a Mean Average Precision (mAP) of 43.09\% and 79.92\% on SYSU-MM01 and RegDB datasets, respectively.
\end{itemize}

\section{Related Work}
\label{sec2}

\subsection{RGB Re-ID}
\label{sec21}

The grateful appearances-based RGB Re-ID approaches~\cite{RN192} mainly focus on how to better learn the semantic representation of high-level features (such as attributes and depth features) or low-level ones (such as shape, color, and texture) that are more discriminative. Along with rapid development of convolutional neural networks, deep-based solutions have achieved promising performance recently. Person Re-ID approaches combined with supervised deep learning can usually be divided into two categories: One is the method based on representation learning, the other is the method based on measurement learning. In the representation-based learning approach, person Re-ID is usually considered as a visual classification problem and the similarity is calculated by using the embedding features projected to the common feature representation space. The researchers hope that these embedding features existing in common space can better describe person images after using human labels in the training stage. Recent works include part based~\cite{RN194,RN195}, GAN based~\cite{RN197,RN198}, and attention based~\cite{RN149,RN193} ones, etc. 

The metric learning based methods can make the model better learn the potential correlation between the data, which is beneficial to the model to learn more discriminant features. The metric learning steered methods aim at learning discriminative features by preform the feature distances comparison between distinguishing samples. Some popular metric learning steered methods~\cite{RN199,RN200,zhao2020deep,zhao2020similarity} proposed well-designed loss functions like triplet loss and contrastive loss to reduce the distances of features from the same identity and meanwhile magnify those from different identities. Angular constraints, also named cosine softmax loss~\cite{RN201}, have been well studied in RGB Re-ID community. The cosine softmax loss is mainly designed by combining L2 normalization with complex classification loss based on softmax to achieve the purpose of enhancing the angle of embedded features and making it easier to distinguish. On the contrary, our proposed approach focuses on designing a more concise ranking loss that directly constrain the angle of the resulting embedded features.

\subsection{RGB-IR Re-ID}
\label{sec22}

For the demand of security in the dark, although infrared camera has been an important part of visual information acquisition, there are still few other studies on cross-modality RGB-IR Re-ID. Recently, to better study cross-modality RGB-IR Re-ID, the researchers proposed a large-scale dataset called SYSU-MM01~\cite{RN26}. In such dataset, the query set images are from the infrared camera, while the gallery set images are from the visible camera, which is more realistic.

Most recent works on cross-modality RGB-IR Re-ID can be divided into two categories: single-stream-based and two-stream-based networks. By padding the multi-modality inputs into domain-specific nodes, Wu et al.~\cite{RN26} proposed a deep zero-padding strategy based single-stream model to match cross-modality features. Wang et al.~\cite{RN48} and Dai et al.~\cite{2018Cross} considered the adversarial training scheme, and designed a special pipeline to guide model learning and achieved remarkable performance. In the related studies based on the two-stream models, HCML~\cite{RN169} and BDTR~\cite{RN188} combined the representation learning and measurement learning simultaneously to constrain the model. On the basis of the two-stream network, MSR ~\cite{RN172} integrated the view classification~\cite{RN202} scheme and proposed a new network to learn the modality speciality and modality sharing features.

\subsection{Cross-Modality Feature Learning}
\label{sec23}

Cross-modal retrieval matches the modalities of input queries with the output results from different modalities. The main goal of cross-modal retrieval is to mitigate the “modality gap” caused by the inconsistent feature representations between different modalities~\cite{gao2021constructing,serikawa2014underwater}. The multilayer deep neural networks (DNNs) based methods have been widely designed to project the heterogeneous features into a consistent semantic space where the feature metric learning can be performed.

For example, Xu et al.~\cite{xu2020cross} proposed cross-modal attention with semantic consistency to perform cross-modality feature embedding for image–text matching task. For effective audio-visual association, Zhang et al.~\cite{zhang2021enhancing} designed a self-supervised curriculum learning method in terms of the teacher-student learning framework. The contrastive learning scheme is explored to distill and capture the cross-modal correspondence. Lu et al.~\cite{lu2020cross} presented a cross-modality shared-specific feature transfer algorithm, which can perform discriminative and complementary feature learning. Ye et al.~\cite{ye2020visible} devised a Homogeneous Augmented Tri-Modal learning method, which performs tri-modal feature learning to reduce cross-modality variations. Wei et al.~\cite{wei2021flexible} proposed an adversarial learning-based flexible body partition model to alleviate the cross-modality gap and promote the feature representation capability.

\section{Methods}
\label{sec3}

\begin{figure*}[t]
	\centering
	\includegraphics[width=4.7in]{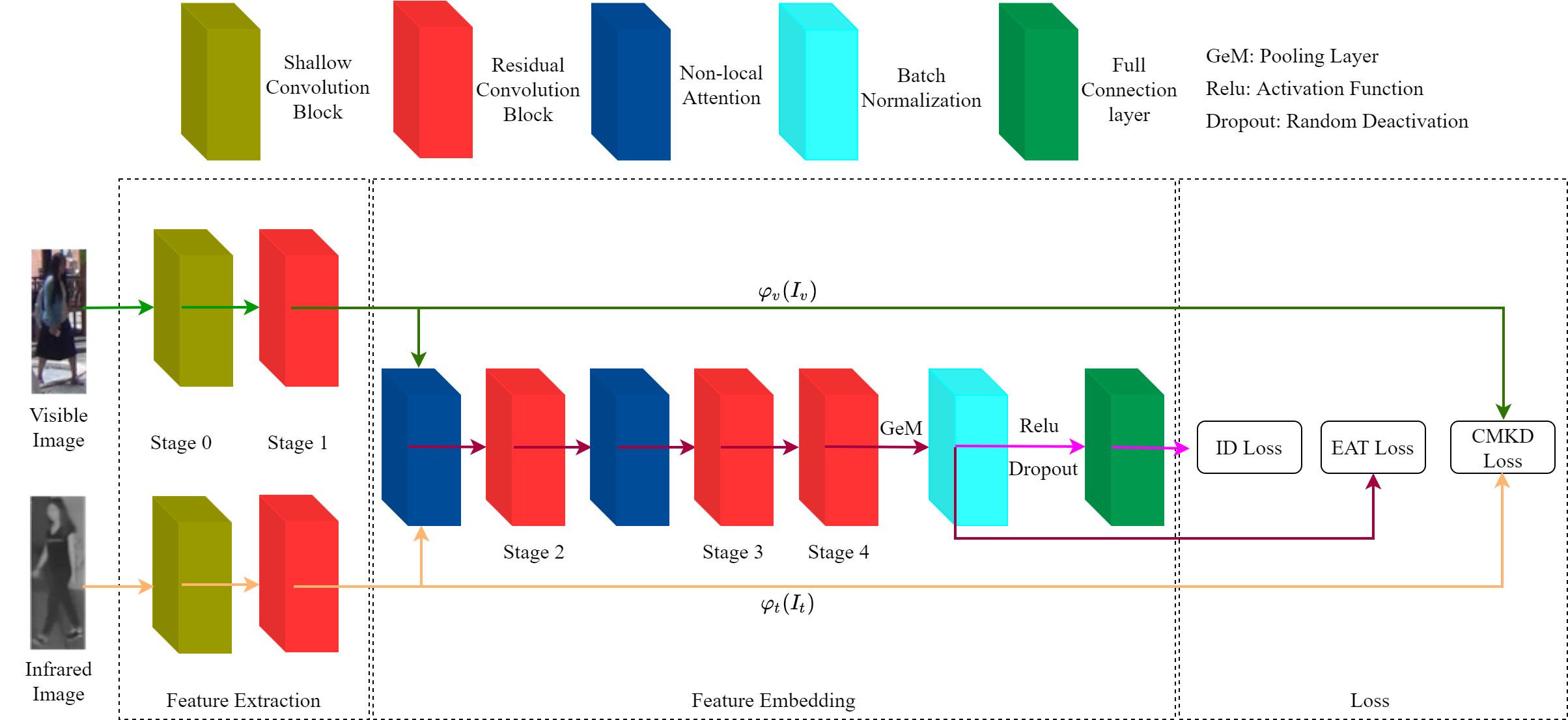}
	\caption{The pipeline of our proposed method for cross-modality person Re-ID. The network mainly consists of two stages. In the first stage, the feature extraction part is composed of two independent branches that do not share parameters, which is used to learn the unique features of the two modes. In the second stage, the feature mapping part is composed of a common network with shared parameters, which is used to map the learned features to a public space. The experiment has three constraints: (1). The proposed Enumerate Angular Triplet (EAT) loss; (2). The proposed Knowledge Distillation (KD) loss; (3). Identity (ID) Loss.}
	\label{Figure 1}
\end{figure*}

In this part, we will detail our presented method for cross-modality person re-identification, as elaborated in Fig.~\ref{Figure 1}. The following description mainly includes three parts: (1) the backbone network, (2) the Enumerate Angular Triplet (EAT) loss, and (3) the Cross-modality Knowledge Distillation (CMKD) loss.

\subsection{Network Backbone}
\label{sec31}

The most commonly used method in the field of visible-thermal cross-modality person re-identification problem is a two-stream network, which was first introduced in~\cite{RN188}. The network is composed of feature extraction and feature embedding. The purpose of the feature extraction is to learn modality-specific information of visible and infrared modality, while the target of the feature embedding is to learn the modality-shared common features between the above two modalities.

In the existing work, the first part (feature extraction) is often implemented directly through some designed convolution neural networks, such as ResNet50, etc, while the second part (feature embedding) is usually implemented by some commonly shared fully connected layers. Feature extraction is mainly composed of two branches that do not share the parameters. But if each branch contains a whole CNN architecture, the number of network parameters can be multiplied. Feature embedding is mainly composed of several full-linked layers with shared parameters. But it can only deal with the 1D-shaped feature vectors and ignore critical personal space structure information about a person.

To handle the above two issues, we take full advantage of the previous experience and divide the convectional CNN model into two parts. ResNet50 is mainly composed of the shallow convolution block stage0 and those res-convolution blocks stage1, stage2, stage3, and stage4. We use stage0 and stage1 as the feature extraction part and stage2, stage3, and stage4 as the next feature embedding part.

For the sake of description, we use ${\varphi _v}$ and ${\varphi _t}$ to represent visible lightweight feature extraction function and infrared feature extraction function, respectively. The above two networks are used to attain modality-specific information. The following feature embedding network is represented by ${\varphi _{vt}}$, which can project the learned features into a common feature representation space. When acquired a visible image ${I_v}$ and an infrared image ${I_t}$, the final feature learned in the common space is described as

\begin{equation}
{V  = }{\varphi _{vt}}{\rm{(}}{\varphi _v}{\rm{(}}{{{I}}_{v}}{\rm{))}}.
\label{eq1}
\end{equation}
\begin{equation}
{T  = }{\varphi _{vt}}{\rm{(}}{\varphi _t}{\rm{(}}{{{I}}_{t}}{\rm{))}}.
\label{eq2}
\end{equation}

The attention scheme has been proven that they can play a vital role in cross-modality Re-ID tasks. We leverage the simple yet effective non-local attention block in~\cite{RN178} to attain the more meaningful descriptions of all positional features, which is represented by

\begin{equation}
z_i = {W_{z}}{\rm{*}}\varphi (x_i) + x_i,
\label{eq3}
\end{equation}
where $\varphi (.)$ denotes a non-local operation, ${W_{z}}$ is the desired weight matrix to be learned and $+ xi$ formulates a residual learning scheme. 

In the person re-identification tasks, neither the most frequently used average pool operation nor the maximum pool operation can capture the domain-specific distinguishing features. Therefore, we utilize a generalized-men (GeM)~\cite{RN2,RN1} pooling layer. The feature mapping of 3D parts is transformed into the feature vector of 1D parts. Given an intermediate 3D feature description $X \in {R^{C \times W \times H}}$, the popular GeM can be represented as

\begin{equation}
{X} = {(\frac{1}{{{|X|}}}\sum\limits_{x_i \in X} {{x^p}_i} )^{\frac{1}{p}}},
\label{eq4}
\end{equation}
where $X \in {R^{C \times 1 \times 1}}$ denotes the desired pooled results, $|.|$ is the element amount, $p$ denotes the pooling hyper-parameter, that can be pre-set or learned by the back-propagating. When ${p} \to \infty $, GeM approximates max-pooling. While when ${p} \to 1$, GeM  approximates  average-pooling.

\subsection{ Cross-Modality Knowledge Distillation Loss}
\label{sec32}

\begin{figure}[t]
	\centering
	\includegraphics[width=3.5in]{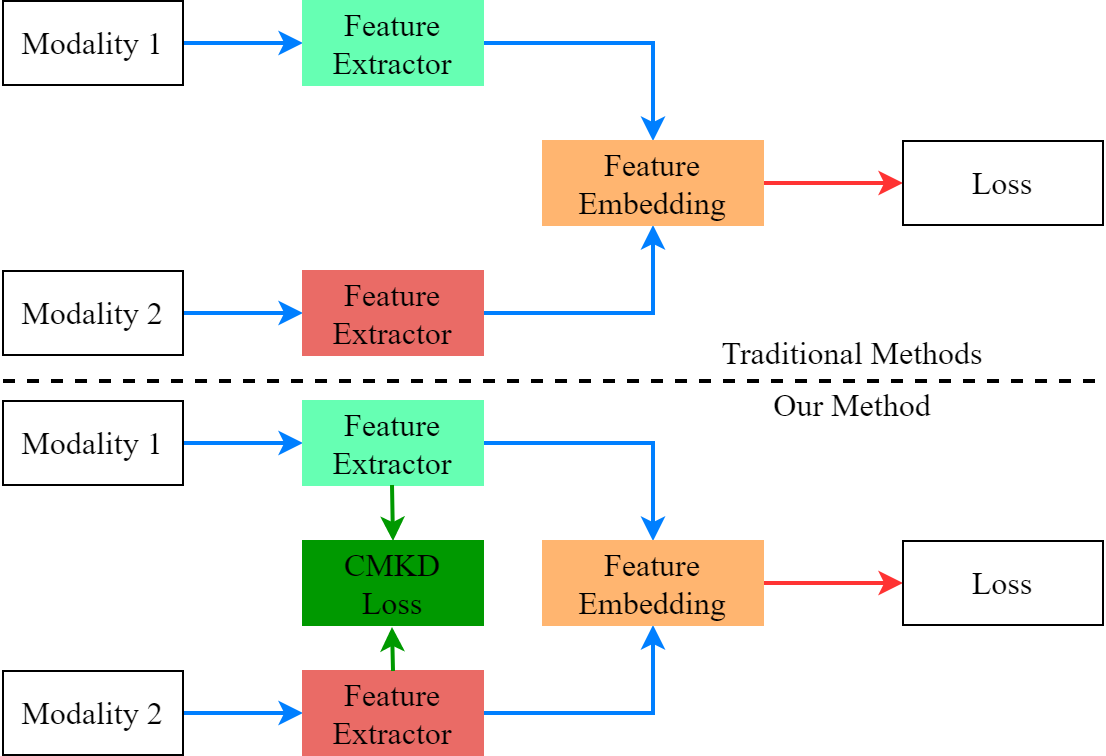}
	\caption{Comparisons between the conventional networks and our designed network. Through our experiments, we found that CMKD loss is beneficial to obtain more discriminant features in the feature embedding space.}
	\label{Figure 2}
\end{figure}

First, we consider narrowing the distance between different modal features in the feature extraction stage to make the following feature embedding more reasonable and effective. At present, most previous methods first learn the unique features of the modal in the feature extraction stage and then map the features between different modalities into a common space after the feature embedding stage. But they have overlooked an important issue, the insurmountable gap between modalities still exists. Therefore, in the feature extraction stage, extracting the unique features of the modality while reducing the distance between the modalities is beneficial to the subsequent work of the feature embedding stage. To this end, as shown in Figure.~\ref{Figure 2}, in the feature extraction stage, we propose a novel loss function, called Cross-Modality Knowledge Distillation (CMKD) loss as follows:

\begin{equation} 
\begin{split} 
L{}_{CMKD} = \frac{1}{N}{\sum\limits_{I_a,I_p} {{{||{V^{rgb}}_a - {V^{ir}}_p||}_2}^2 + 
{{||{V^{ir}}_a-{V^{rgb}}_p||}_2}} ^2},
\end{split} 
\label{eq5}
\end{equation}
where ${V^{rgb}}_a$ and ${V^{rgb}}_p$ is the embedding features of anchor sample and positive sample in RGB modality, ${V^{ir}}_a$ and ${V^{ir}}_p$ is the embedding features of anchor sample and positive sample in infrared modality, and ${||.||}_2$ denotes the Euclidean distance.

\subsection{Enumerate Angular Triplet Loss}
\label{sec33}

The triple loss function is designed to make the distance between the anchor image ${I_a}$ and the positive image ${I_p}$ closer than that between the anchor sample ${I_a}$ and the negative sample${I_n}$ by a constraint $\zeta $.

 As the triplet loss is difficult to separate the orientation of the embedding features, we put forward a novel bi-directional Enumerate Angular Triplet (EAT) loss function. Our proposed loss function uses not Euclidean distance but cosine distance to represent the similarities between embedded features.
 
Taking those above factors into account, we first design an efficient loss function named angular triple (AT) loss as follows:

\begin{equation}
{L_{\cos }} = \frac{1}{N}\sum\limits_{I_a,I_p,I_n} {{{[\cos ({V_a},{V_n}) - \cos ({V_a},{V_p}) + \zeta ]}_ + }},
\label{eq6}
\end{equation}
where $N$ is the number of the identity class, $V_a$ is the embedding features of the anchor sample ${I_a}$, $V_n$ is the embedding features of the negative sample ${I_n}$, and $V_p$ is the embedding features of the positive sample ${I_p}$.

However, the proposed function suffers from two drawbacks. First, it is our target to promote ${{V_a}}$ and ${{V_n}}$ easier to be distinguished in the embedded feature representation space. Therefore, there needs to be a clamping function to achieve the above purpose. Second, the overall clamping function is excluded with an appropriate selection of margin $\zeta $. When the margin is not less than 1, the loss function is usually non-negative. Therefore, we set the margin to be 1 to maintain the parameters uncomplicated.

Given the above considerations, we then reformulate the AT loss function as follows:

\begin{equation}
{L_{AT}} = \frac{1}{N}\sum\limits_{I_a,I_p,I_n} {({{[\cos ({V_a},{V_n})]}_ + } - {{[\cos ({V_a},{V_p})]}_ + } + 1)}.
\label{eq7}
\end{equation}

Regarding the cross-modality enumeration loss function, we first consider the inter-class cross-modality constraints ${L_{crgb}}$ and ${L_{cir}}$, which are similar to the direct triplet loss:

\begin{equation} 
\begin{split} 
{L_{crgb}}  &= \frac{1}{N}\sum\limits_{I_a,I_p,I_n} \{({{[\cos ({V^{rgb}}_a,{V^{ir}}_n)]}_ + }  \\ 
& - {{[\cos ({V^{rgb}}_a,{V^{ir}}_p)]}_ + } + 1 \}, 
\end{split}
\label{eq8}
\end{equation}

\begin{equation} 
\begin{split} 
{L_{cir}}  &= \frac{1}{N}\sum\limits_{I_a,I_p,I_n} \{({{[\cos ({V^{ir}}_a,{V^{rgb}}_n)]}_ + }  \\ 
& - {{[\cos ({V^{ir}}_a,{V^{rgb}}_p)]}_ + } + 1 \}. 
\end{split} 
\label{eq9}
\end{equation}

Based on the above formulations, the inter-class same-modality constraints ${L_{srgb}}$ and ${L_{sir}}$ are designed to mitigate the modality gap at the image patch level:

\begin{equation} 
\begin{split} 
{L_{srgb}}  &= \frac{1}{N}\sum\limits_{I_a,I_p,I_n} \{({{[\cos ({V^{rgb}}_a,{V^{rgb}}_n)]}_ + }  \\ 
& - {{[\cos ({V^{rgb}}_a,{V^{ir}}_p)]}_ + } + 1 \},
\end{split} 
\label{eq10}
\end{equation}

\begin{equation} 
\begin{split} 
{L_{sir}}  &= \frac{1}{N}\sum\limits_{I_a,I_p,I_n} \{({{[\cos ({V^{ir}}_a,{V^{ir}}_n)]}_ + }  \\ 
& - {{[\cos ({V^{ir}}_a,{V^{rgb}}_p)]}_ + } + 1 \}. 
\end{split} 
\label{eq11}
\end{equation}

In summary, we can formulate the bi-directional extension of AT loss as follows:

\begin{equation}
{L_{ATrgb}} = {L_{crgb}} + {L_{srgb}},
\label{eq12}
\end{equation}

\begin{equation}
{L_{ATir}} = {L_{cir}} + {L_{sir}},
\label{eq13}
\end{equation}

\begin{equation}
{L}_{EAT} = {L_{ATrgb}} + {L_{ATir}}.
\label{eq14}
\end{equation}

\begin{figure*}[t]   
\centering
\includegraphics[width=2.3in,trim=18 15 6 5]{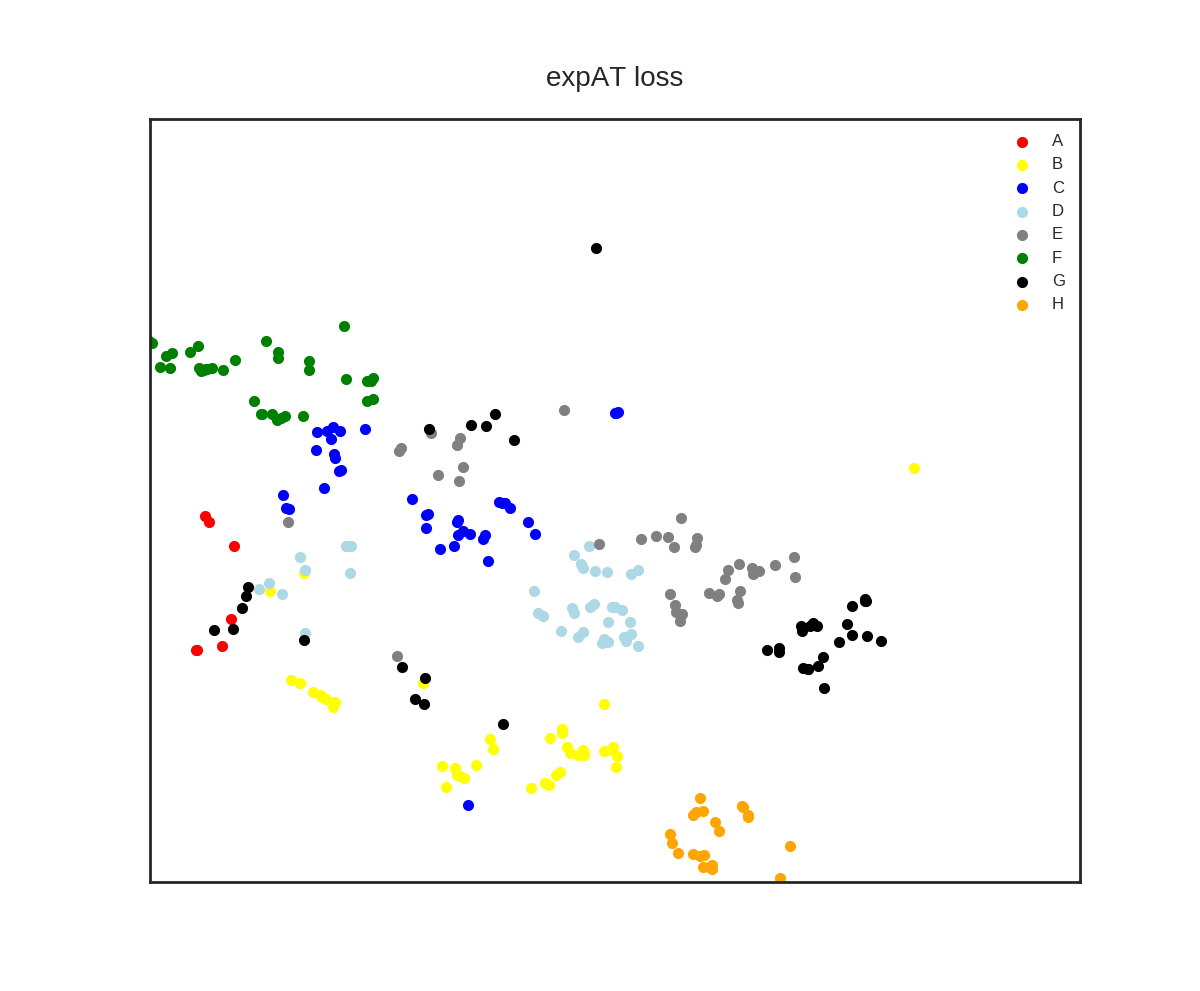}
\includegraphics[width=2.3in,trim=18 15 6 5]{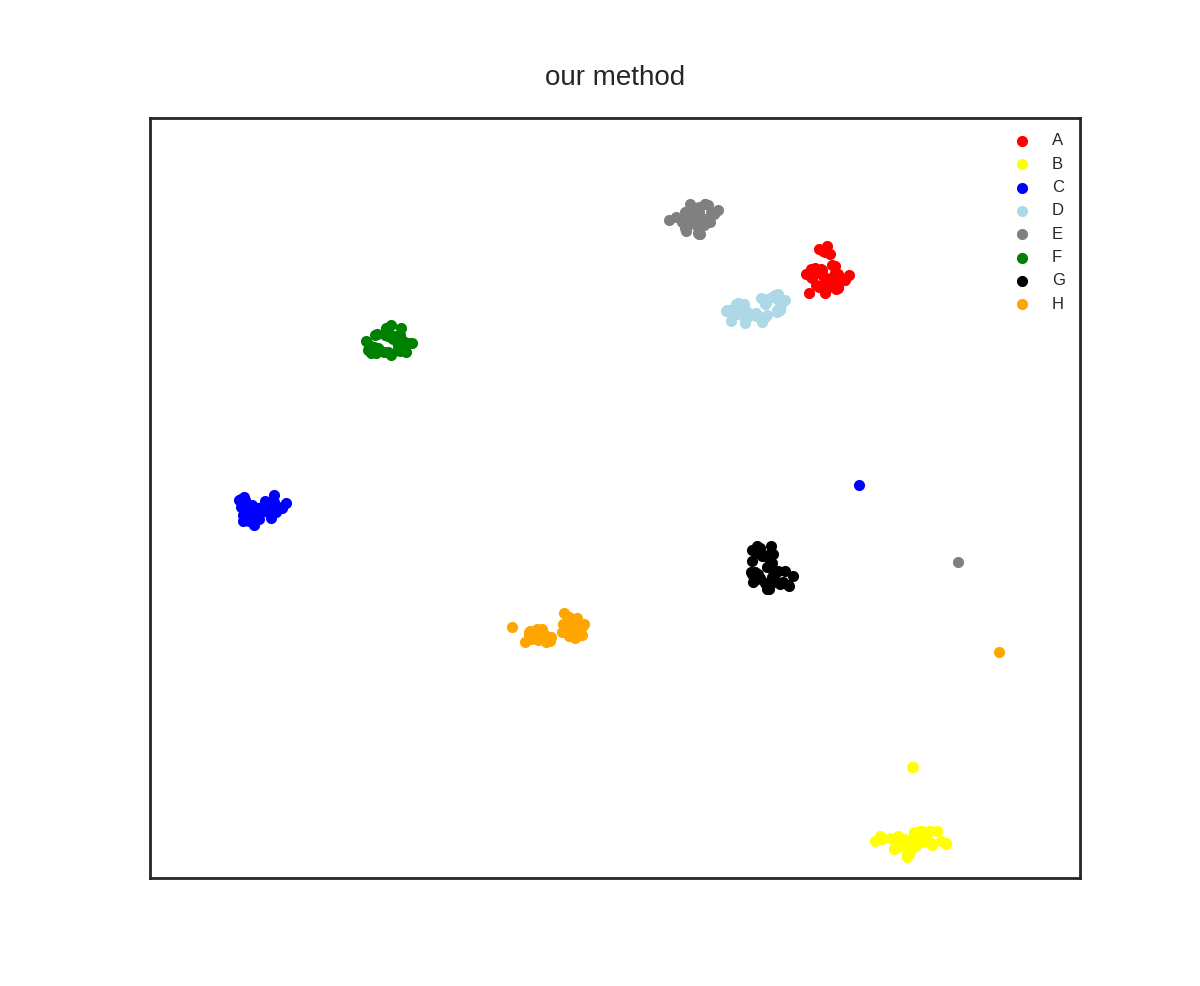}
\caption{Two-dimensional visualization of query subset features in the SYSU-MM01 dataset in the common feature space. Compared with expAT loss~\cite{RN189}, the embedded vectors of different classes in our proposed method are easier to be separated in the feature space.}
\label{Figure 3}
\end{figure*}

Moreover, since the exponential function $y = {e^x}$ grows exponentially at  $x > 0$, this characteristic is beneficial to the rapid convergence of the model. Thus, we give the bi-directional enumerate AT loss as

\begin{equation}
\begin{array}{l}
{L_{EATrgb}} = \frac{1}{N}\sum\limits_{I_a,I_p,I_n} {{e^{({{[\cos ({V^{rgb}}_a,{V^{ir}}_n)]}_ + } - {{[\cos ({V^{rgb}}_a,{V^{ir}}_p)]}_ + } + 1)}}} \\
+ \frac{1}{N}\sum\limits_{I_a,I_p,I_n} {{e^{({{[\cos ({V^{rgb}}_a,{V^{rgb}}_n)]}_ + } - {{[\cos ({V^{rgb}}_a,{V^{ir}}_p)]}_ + } + 1)}}}, 
\end{array}
\label{eq15}
\end{equation}

\begin{equation}
\begin{array}{l}
{L_{EATir}} = \frac{1}{N}\sum\limits_{I_a,I_p,I_n} {{e^{({{[\cos ({V^{ir}}_a,{V^{rgb}}_n)]}_ + } - {{[\cos ({V^{ir}}_a,{V^{rgb}}_p)]}_ + } + 1)}}} \\
+ \frac{1}{N}\sum\limits_{I_a,I_p,I_n} {{e^{({{[\cos ({V^{ir}}_a,{V^{ir}}_n)]}_ + } - {{[\cos ({V^{ir}}_a,{V^{rgb}}_p)]}_ + } + 1)}}}, 
\end{array}
\label{eq16}
\end{equation}

\begin{equation}
{L}_{EAT} = {L_{EATrgb}} + {L_{EATir}}.
\label{eq17}
\end{equation}

However, in our experiments, we find that loss function~(\ref{eq17}) is difficult to be converged. In our designed enumeration loss, to ensure that each element of the generated feature description is as evenly distributed as possible, thus making the descriptor more informative and discriminative, a compactness term $C$ is introduced. It is computed by comparing the differences between each element in $f(.)$ and the mean value:

\begin{equation} 
\begin{split} 
C &= \sum\limits_i^N {\sum\limits_r^R {{e^{\cos (fr({V^{rgb}}_a),\overline f ({V^{rgb}}_a))}}} }  + \\ &\sum\limits_i^N {\sum\limits_r^R {{e^{\cos (fr({V^{ir}}_a),\overline f ({V^{ir}}_a))}}} }. 
\end{split} 
\label{eq18}
\end{equation}

Here, ${f_r}(.)$ and $\overline f (.)$ denotes the $r$-th element and the mean index of all elements in $f(.)$, respectively. $R$ denotes the length of the acquired local deep descriptor $f(.)$. The introduction of $C$ can also avoid overfitting in network training. In our experiment, if the compactness term in the loss is not enumerated, it is difficult for the network to converge.

In summary, our enumerate angular triplet loss function is defined by integrating them as follows:

\begin{equation}
{L_{EAT}} = {L_{EAT}} + C.
\label{eq19}
\end{equation}

Fig.~\ref{Figure 3} plots the two-dimensional visualization of the features in the common feature space with respective methods. We can observe that by using our proposed EAT loss which considers both intra-modal and inter-modal distance, the embedding features from different classes in the common feature space are drastically separable, which contributes to the following classification performance. On the contrary, the expAT loss only considers the distance between images from different modalities and ignores the intra-modal distance that can provide discriminative and complementary information.

\subsection{Overall Loss}
\label{sec34}

In addition, to achieve better classification results, similar to some of the advanced methods, we also take into account identity loss, which integrates identity-specific information by treating each person as one class. The formula for identity loss is given as follows:

\begin{equation} 
\begin{split} 
{L_{{ID}}} = \sum\limits_{i = 1}^N { - {q_i}\log ({p_i})}, 
\end{split} 
\label{eq20}
\end{equation}
where ${p_i}$ is the  prediction label of the ${i^{th}}$ class, ${q_i}$ is the true label of the ${i^{th}}$ class, $N$ is the number of all classes of the training samples.

So far, the final loss of our method is calculated as follows:

\begin{equation} 
\begin{split} 
{L_{ALL}} = {L_{EAT}} + {L_{CMKD}} + {L_{ID}}.
\end{split} 
\label{eq21}
\end{equation}

\begin{figure}[t]
	\centering
	\includegraphics[width=3.4in]{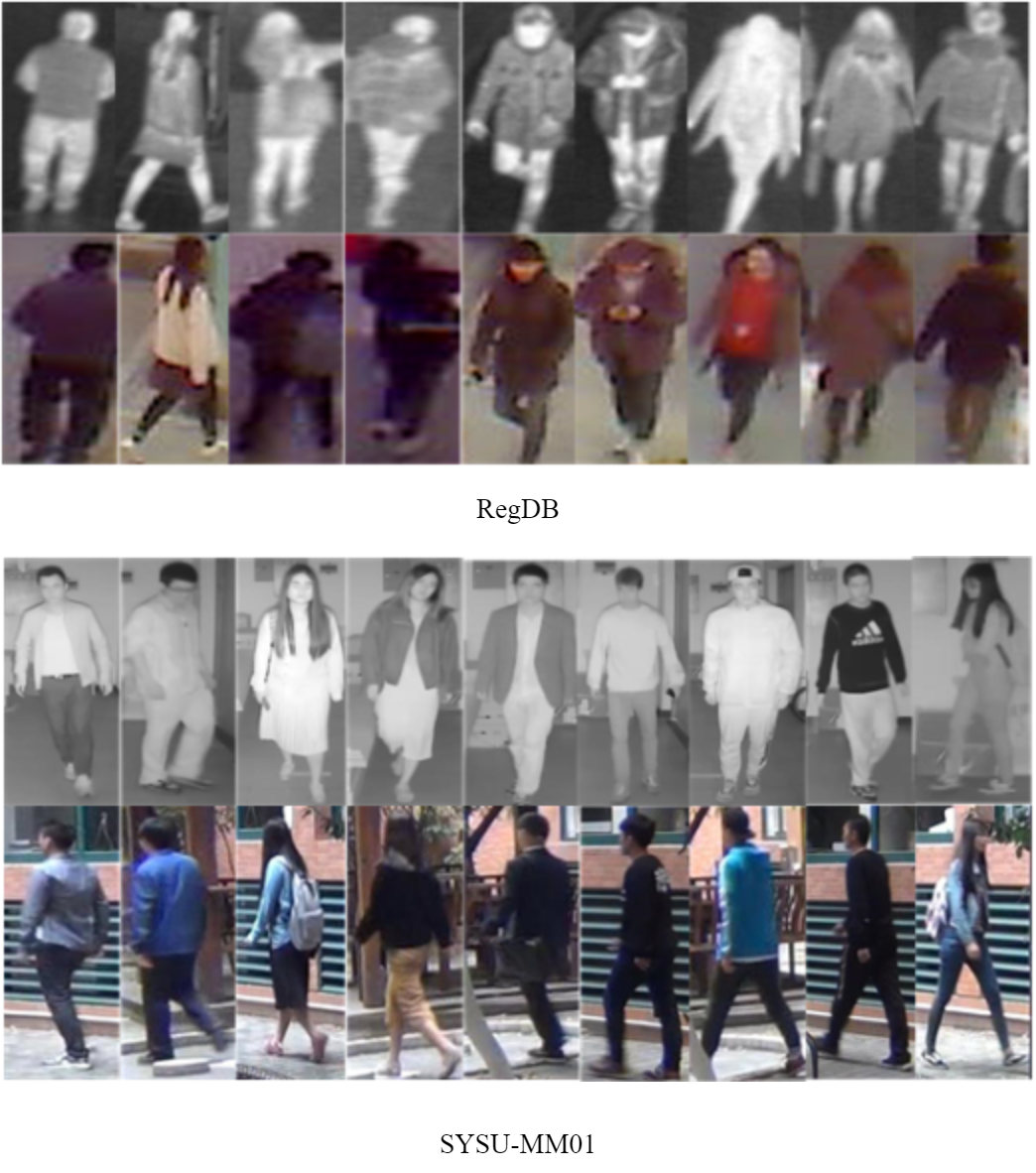}
	\caption{Examples of cross-modality images. The images in the first part are from the RegDB set, where the first row shows the images captured by a thermal imaging camera and the second row shows the images captured by a visible light camera. The images in the second part are from the SYSU-MM01 set, where the first row shows the images captured by the infrared camera and the second row shows the image captured by the visible light camera. Two images in each column from the same identity.}
	\label{Figure 4}
\end{figure}

\section{Experiments}
\label{sec4}

In this section, we conducted evaluations on RegDB~\cite{RN51} (providing infrared images by thermal cameras) and SYSU-MM01~\cite{RN26} (providing infrared images by near-infrared cameras) datasets to verify the efficiency and effectiveness of our proposed method. The example images are listed in Fig.~\ref{Figure 4}.

\subsection{Datasets}
\label{sec41}

1) SYSU-MM01 set: The SYSU-MM01 dataset is a dataset consisting of 491 identities providing RGB and IR images from six cameras. The training set contains 395 people, including visible images 22,258 pieces and infrared images 11909 pieces. The other 96 people are included in the testing set, including 3803 infrared samples for query and 301 randomly selected visible samples as the gallery set.

2) RegDB set: The RegDB dataset is one of the most commonly used datasets in the RGB-thermal person re-identification field. It consists of 8,240 photos from 412 identities, each with 10 visible images and 10 infrared images.

\subsection{Evaluation Metrics}
\label{sec42}

Based on previous work, Cumulative Matching Characteristics (CMC) and Mean Average Precision (mAP) are used as the evaluation indexes. To evaluate the accuracy of the results, we repeated the final evaluation 10 times, randomly segmented each time with different gallery sets, and calculated statistically stable performance indicators as the final results.

\subsection{Implementation Details}
\label{sec43}

Experiments are carried out on two datasets to verify the proposed method, and the algorithm is implemented using Pytorch 1.2. The batch size is set to 8 and the proposed method is optimized using the ADAM optimizer ~\cite{RN187} with a warm-up ~\cite{RN186} strategy, with an initial learning rate set as $3 \times 10^{-4}$ and the decay of 0.1 at 10,000 and 20,000 steps. The training process is iterated by 30,000 steps. For the anchor of both modalities, the margin of the triplet loss is empirically set as 0.3. We also use the Label Smoothing~\cite{RN206} and the Random Erasing~\cite{RN183} strategies to alleviate the overfitting problem in the training stage, and the corresponding parameters are set as 0.1 and 0.5 respectively. In the test, Euclidean distance is used to calculate the feature distance.

Batch Sampling Strategy: To better perform cross-modality constraints, we use a special sampling strategy. First, we use a visible sample and an infrared sample of the same identity as an image pair. For the batch size $N$, we randomly select $N$ anchor sample pairs from the entire training set. Then, a negative RGB sample and a positive RGB sample are randomly selected from the training dataset where the RGB images of anchor points are eliminated, to form a tuple with the anchor. So there's a total of $N \times 2 \times 3$ samples in each mini-batch. In our experiment, anchor samples of the same identity from different modalities of the same tuple are explored to calculate the identity loss, and all images in the same tuple are explored to calculate the ranking loss. We use the random sampling strategy to traverse all the training samples in the training process.

\begin{table}[t]
	\caption{ A{\upshape blation studies (\%) of the presented method on SYSU-MM01 set.} W{\upshape e study the effects of three different component combinations including EAT loss, non-local attention and CMKD loss.}}
	\begin{center}
		\setlength\tabcolsep{5pt}
		{\fontsize{7pt}{9pt}\selectfont
			
				\begin{tabular}{|c|c|c|c|c|c|c|}
				\hline
				\multirow{2}{*}{\textbf{EAT Loss}}&\multirow{2}{*}{\textbf{Non-Local Attention}}	&\multirow{2}{*}{\textbf{CMKD Loss}}&\multicolumn{4}{|c|}{\textbf{SYSU-MM01}}\\
				\cline{4-7}  
				& & &\textbf{\textit{Rank1}} &\textbf{\textit{Rank10}} &\textbf{\textit{Rank20}} &\textbf{\textit{mAP}}\\
				\hline
				\XSolid &\XSolid 		&\XSolid 	&21.14 	&50.25 	&62.77 	&18.52  \\
				\hline
				\Checkmark &\XSolid 		&\XSolid 	&40.17 	&78.73 	&87.64 	&40.11  \\
				\hline
				\Checkmark &\Checkmark		&\XSolid	&42.16	&78.76	&87.52  &41.88  \\
				\hline
				\Checkmark &\XSolid		    &\Checkmark	&41.80	&80.07	&89.26  &41.29  \\
				\hline
				\Checkmark &\Checkmark		&\Checkmark	&43.23	&82.78	&90.91  &43.09  \\
				\hline
			\end{tabular}}
		\label{tab1}
	\end{center}
\end{table}

\begin{table}[t]
	\caption{ C{\upshape omparison performance (\%) on the SYSU-MM01 set.}  “-”{\upshape indicates not available or not provided.}}
	\begin{center}
		\setlength\tabcolsep{5pt}
		{\fontsize{7pt}{9pt}\selectfont
			
			\begin{tabular}{|c|c|c|c|c|c|c|c|c|}
				\hline
				\textbf{}&\multicolumn{4}{|c|}{\textbf{All Search}} &\multicolumn{4}{|c|}{\textbf{Indoor Search}}\\
				\hline
				\textbf{Methods} & \textbf{\textit{Rank1}}& \textbf{\textit{Rank10}}& \textbf{\textit{Rank20}} & \textbf{\textit{mAP}}& \textbf{\textit{Rank1}}& \textbf{\textit{Rank10}}& \textbf{\textit{Rank20}} & \textbf{\textit{mAP}}\\
				\hline
				Zero-pad~\cite{RN26}  &14.80		&54.12	&71.33 	&15.95   &20.58		&68.38	&85.79 	&26.92    \\
				\hline
				cmGAN~\cite{2018Cross}     &26.97	&67.51	&80.56 	&27.80    &31.63		&77.23	&89.18 	&42.19    \\
				\hline
				HCML~\cite{RN169}  	  &14.32	&53.16	&69.17 	&16.16   &24.52		&73.25	&86.73 	&30.08    \\
				\hline
				HSME~\cite{RN167}  	  &20.68	&62.74	&77.95 	&23.12   &-		    &-	    &- 	    &-    \\
				\hline
				$\mathrm{D}^{2} \mathrm{RL}$~\cite{RN48}      &28.90		&70.60	&82.40 	&29.20    &-		    &-	    &- 	    &-    \\
				\hline
				MAC~\cite{RN166} 	  &33.26	&79.04	&90.09 	&36.22   &36.43		&62.36	&71.63 	&37.03    \\
				\hline
				AlignGAN~\cite{RN140}  &42.40		&85.00	    &93.70 	&40.70    &45.90		&87.60	&94.40 	&54.30    \\
				\hline
				HPLIN~\cite{RN151}     &41.36	&84.78	&94.51 	&42.95   &45.77		&91.82	&98.46 	&56.52    \\
				\hline
				Hi-CMD~\cite{RN170}    &34.94	&77.58	&- 	    &35.94   &-		    &-	    &- 	    &-    \\
				\hline
				EDFL~\cite{RN163}      &36.94	&85.42	&93.22 	&40.77   &-			&-		&- 		&-    \\
				\hline
				CDP~\cite{RN180}  	  &38.00		&82.30	&91.70 	&38.40    &-			&-		&- 		&-    \\
				\hline
				expAT~\cite{RN189}  	  &38.57	&76.64	&86.39 	&38.61   &-			&-		&- 		&-    \\
				\hline
				eBDTR~\cite{8732420}     &27.82	&67.34	&81.34 	&28.42   &32.46		&77.42	&89.62 	&42.46    \\
				\hline
				MSR~\cite{RN172}  	  &37.35	&83.40	&93.34 	&38.11   &39.64		&89.29	&97.66 	&50.88    \\
				\hline
				JSIA~\cite{RN181}      &38.10		&80.70	&89.90 	&36.90    &43.80		&86.20	&94.20 	&52.90    \\
				\hline
				Ours      &\textbf{43.23}	&82.78	&90.91 	&\textbf{43.09}   &\textbf{50.07}		&90.63	&96.99 	&\textbf{58.88}    \\
				\hline
		\end{tabular}}
		\label{tab2}
	\end{center}
\end{table}

\begin{table}[t]
	\caption{ P{\upshape eer comparisons (\%) on the RegDB set under “Thermal to RGB” and “RGB to Thermal” experimental settings.}  }
	\begin{center}
		\setlength\tabcolsep{14pt}
		{\fontsize{8pt}{9pt}\selectfont
			
			\begin{tabular}{|c|c|c|c|c|}
				\hline
				\textbf{}&\multicolumn{2}{|c|}{\textbf{Thermal to RGB}} &\multicolumn{2}{|c|}{\textbf{RGB to Thermal}}\\
				\hline
				\textbf{Methods} & \textbf{\textit{Rank1}} & \textbf{\textit{mAP}}& \textbf{\textit{Rank1}} & \textbf{\textit{mAP}}\\

				\hline
				Zero-Pad~\cite{RN26}  &16.63	&17.82	   &17.75 	&31.83        \\
				\hline
				HCML~\cite{RN169}      &21.70		&22.24		&24.44 	&20.08      \\
				\hline
				$\mathrm{D}^{2} \mathrm{RL}$~\cite{RN48}      &-		&-			&43.40 	&44.10      \\
				\hline
				MSR~\cite{RN172}       &-		&-			&48.43 	&48.67      \\
				\hline
				AlignGAN~\cite{RN181}  &56.30		&53.40		&- 		&-        \\
				\hline
				EDFL~\cite{RN163}  	  &51.89	&52.13		&52.58 	&52.98      \\
				\hline
				AGW~\cite{RN182}       &-		&-			&70.05 	&66.37      \\
				\hline
				CMSP~\cite{RN208}  	  &-		&-			&65.07 	&64.50     \\
				\hline
				expAT~\cite{RN189}     &67.45	&66.51		&66.48 	&67.31    \\
				\hline
				Hi-CMD~\cite{RN170}     &-	&-		&70.93 	&66.04    \\
				\hline
				cm-SSFT~\cite{lu2020cross}     &71.00	&71.70		&72.30 	&72.90    \\
				\hline
				HAT~\cite{ye2020visible}     &70.02	&66.30		&71.83 	&67.56    \\
				\hline
				FBP-AL~\cite{wei2021flexible}     &70.05	&66.61		&73.98 	&68.24    \\
				\hline
				Ours      &\textbf{80.97} &\textbf{79.92} &\textbf{79.27} &\textbf{77.69}    \\
				\hline
		\end{tabular}}
		\label{tab3}
	\end{center}
\end{table}

\subsection{Ablation Study}
\label{sec44}

In this part, we evaluate and analyze the effectiveness of the proposed method through qualitative and quantitative tests on the popular SYSU-MM01 set. In this experiment, the results are reported with all search single-shot settings. Specifically, “EAT Loss” represents the Enumerate Angular Triplet (EAT) loss, “Non-Local Attention” represents the non-local attention block, and “CMKD Loss” represents the Cross-modality Knowledge Distillation (CMKD) loss. 

We can make the following observations through the results shown in Table~\ref{tab1}. (1) By using the network with EAT loss, we can achieve better performance than the two-stream network in~\cite{RN189,RN172,RN181}. This experiment demonstrates that an angularly informative and discriminative feature space is explicitly beneficial for cross-modality Re-ID. (2) By using CMKD loss, performance is improved by narrowing the distance between different modal features in the feature extraction stage. (3) When aggregating two losses with non-local attention, the performance is further improved, demonstrating that these losses and attention are mutually beneficial to each other.

\begin{figure}[t]  
\centering
\includegraphics[width=3in]{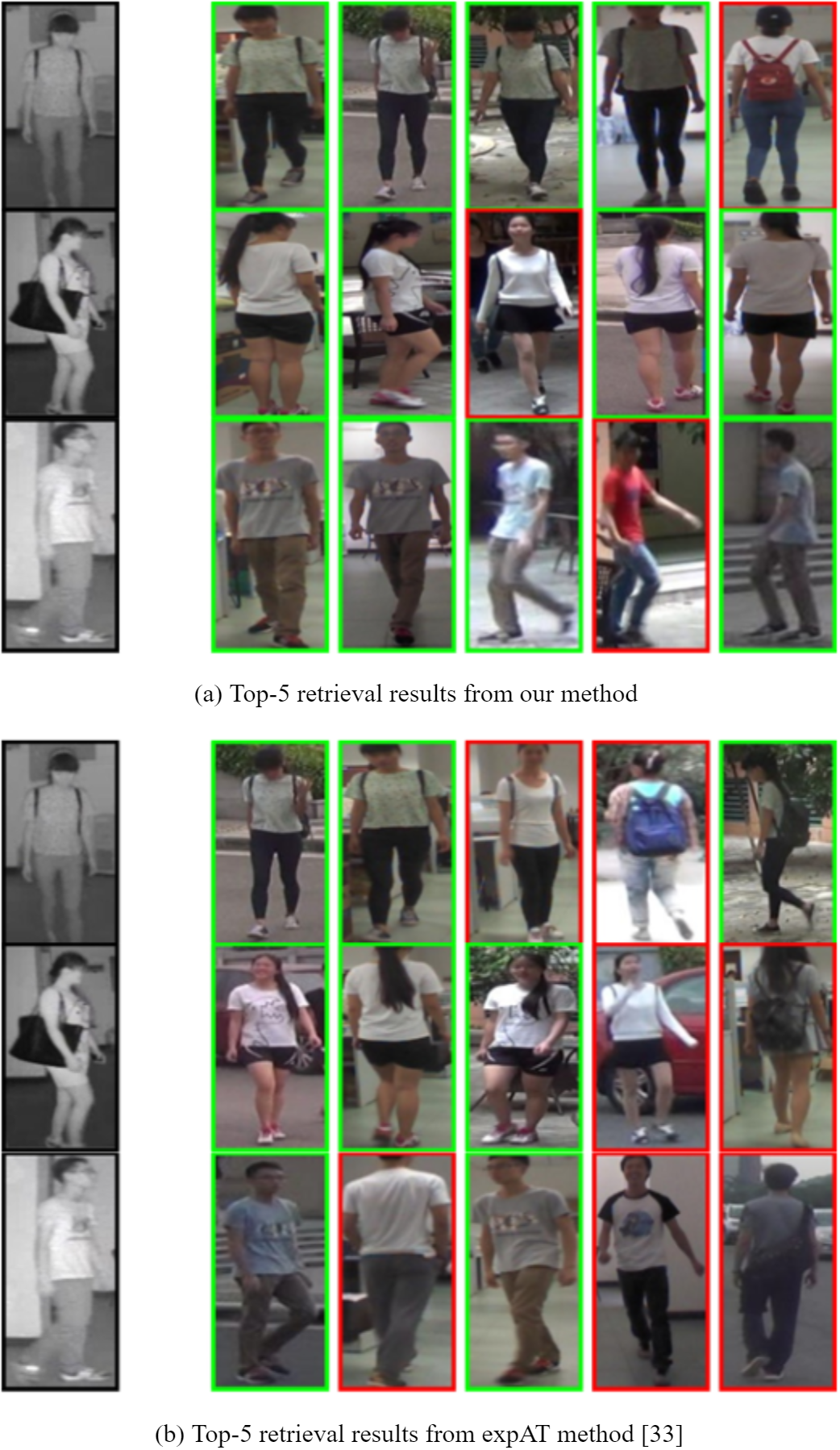}
\caption{Top-5 retrieval results on popular SYSU-MM01 set under “Infrared to RGB” setting. For each query on the left, top-$k$ candidates are listed in ascending order due to their similarities. The false and true retrievals are given in the red and green boxes, respectively.}
\label{Figure 5}
\end{figure}

\begin{figure}[t]  
\centering
\includegraphics[width=3in]{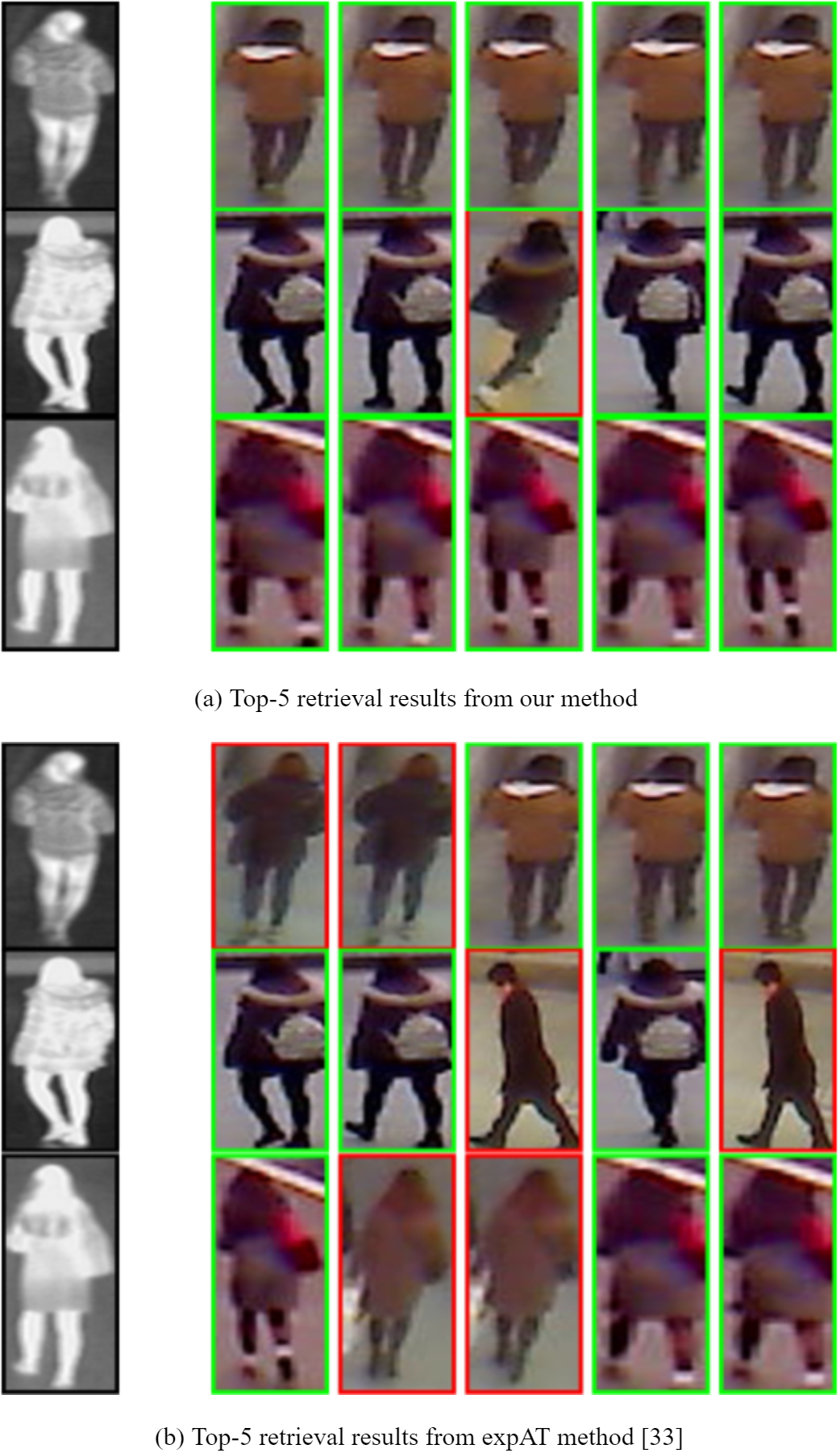}  
\caption{Top-5 retrieval results on popular RegDB dataset under “Thermal to RGB” setting. For each query on the left, top-$k$ candidates are listed in ascending order due to their similarities. The false and true retrievals are given in the red and green boxes, respectively.}
\label{Figure 6}
\end{figure}

\subsection{Peer Comparisons}
\label{sec45}

In this section, we list and compare some state-of-the-art approaches, including eBDTR~\cite{RN182}, HSME~\cite{RN167}, $\mathrm{D}^{2} \mathrm{RL}$~\cite{RN48}, MAC~\cite{RN166}, MSR~\cite{RN172}, AlignGAN~\cite{RN140}, EDFL~\cite{RN163}, and HPILN~\cite{RN151}, etc. Most of the methods we compared were published in the last two years. The results performance on the popular SYSU-MM01 and REGDB sets are provided in Table~\ref{tab2} and Table~\ref{tab3}, respectively.

The experimental results on SYSU-MM01 datasets demonstrate that our proposed method achieves the best performance in all query settings, achieving 43.23${\rm{\% }}$/43.09${\rm{\% }}$ rank-1/mAP for the All Search setting and 50.07${\rm{\% }}$/58.88${\rm{\% }}$ rank-1/mAP for the Indoor Search setting. The experimental results on RegDB dataset also demonstrate that our method achieves the best performance in both query settings, usually by a large margin, achieving 79.27${\rm{\% }}$/77.69${\rm{\% }}$ rank-1/mAP for the RGB to thermal query setting and 80.97${\rm{\% }}$/79.92${\rm{\% }}$ rank-1/mAP for the thermal to RGB query setting.

Also, no additional parameters are introduced into the proposed method during testing, which indicates that this method is easier to use in practical application scenarios. We also provide some retrieval results in Fig.~\ref{Figure 5} and Fig.~\ref{Figure 6}. A large number of above experiments have shown that cross-modal shared feature representation for cross-modal person re-recognition tasks can be better learned by using our proposed EAT loss and CMKD loss.

\section{Conclusions}

The purpose of this paper is to improve  discriminative feature learning by a simple method. On one hand, motivated by the knowledge distillation, a new Cross-Modality Knowledge Distillation (CMKD) loss is explicitly presented to reduce the modality discrepancy in the modality-specific feature extraction stage. On the other hand, in order to help the deep network learn angularly representative embedded features from different modalities, we put forward the Enumerate Angular Triplet (EAT) loss. The EAT loss can constrain the included angle between the embedded vectors, which is helpful for angular segmentation of the feature space. Experimental results on two cross-modality Re-ID datasets have shown that the proposed method is effective compared with the most advanced methods.

%\iffalse
\begin{acknowledgements}
This work was supported in part by the National Key Research and Development Program of China under Project nos. 2018AAA0100102 and 2018AAA0100100, the National Natural Science Foundation of China under Grant nos. 61972212, 61772568 and 61833011, the Natural Science Foundation of Jiangsu Province under Grant no. BK20190089, the Six Talent Peaks Project in Jiangsu Province under Grant no. RJFW-011, Youth science and technology innovation talent of Guangdong Special Support Program, and Open Fund Project of Provincial Key Laboratory for Computer Information Processing Technology (Soochow University) (No. KJS1840).
\end{acknowledgements}
%\fi

\section*{Conflict of interest}

The authors declare that they have no conflict of interest.

\bibliographystyle{spmpsci}      % mathematics and physical sciences
\bibliography{mybibliography}   % name your BibTeX data base

\end{document}